\documentclass[10pt,twocolumn,letterpaper]{article}

\usepackage{iccv}              

\definecolor{iccvblue}{rgb}{0.21,0.49,0.74}
\usepackage[pagebackref,breaklinks,colorlinks,allcolors=iccvblue]{hyperref}
\usepackage{CJKutf8}

\usepackage[utf8]{inputenc} 
\usepackage[T1]{fontenc}    
\usepackage{url}            
\usepackage{nicefrac}       
\usepackage{lipsum}		
\usepackage{doi}

\usepackage{xcolor}         

\usepackage{times}
\usepackage{microtype} %
\usepackage{balance} 

\usepackage{hyperref}

\usepackage{helvet}
\usepackage{courier}

\usepackage{makecell}
\usepackage{graphicx}
\usepackage{color}
\usepackage{amsfonts}
\usepackage{amsmath}
\usepackage{amssymb}

\usepackage{multirow}
\usepackage{booktabs}
\usepackage{wrapfig}

\def\ie{\mbox{\textit{i.e.}, }}
\def\eg{\mbox{\textit{e.g.}, }}

\DeclareMathAlphabet\mathbfcal{OMS}{cmsy}{b}{n}

\def\0{{\bf 0}}
\usepackage{ dsfont }
\def\1{\mathds{1}}

\def\bx{{\bf x}}
\def\by{{\bf y}}

\def\bx{{\bf x}}

\def\by{{\bf y}}

\usepackage{ntheorem}

\newtheorem*{*thm}{Theorem}

\newtheorem*{*lemma}{Lemma}

\usepackage{enumitem}

\newcommand{\yes}{\textcolor[rgb]{0,0,0}{\large{\checkmark}}}
\newcommand{\no}{\textcolor[rgb]{0,0,0}{\large{{\bf\times}}}}

\usepackage{array}
\usepackage{tabu}
\makeatletter
\newcommand{\nipstophline}{%
	\noalign {\ifnum 0=`}\fi \hrule height 4pt
	\futurelet \reserved@a \@xhline
}
\newcommand{\nipsbottomhline}{%
	\noalign {\ifnum 0=`}\fi \hrule height 1pt
	\futurelet \reserved@a \@xhline
}

\def\ljc{\textcolor{black}}

\def\xie{\textcolor{black}}

\def\cam{\textcolor{black}}

\def\lightgray{\textcolor{gray}}

\newcommand\blfootnote[1]{%
  \begingroup
  \renewcommand\thefootnote{}\footnote{#1}%
  \addtocounter{footnote}{-1}%
  \endgroup
}

\def\paperID{***} 
\def\confName{ICCV}
\def\confYear{2025}

\title{LMM-Det: Make Large Multimodal Models Excel in Object Detection}

\author{Jincheng Li\textsuperscript{*}
\quad
Chunyu Xie\textsuperscript{*}
\quad
Ji Ao
\quad
Dawei Leng\textsuperscript{$\dag$}
\quad
Yuhui Yin \\ \\
360 AI Research
}

\begin{document}
\twocolumn[{%
\renewcommand\twocolumn[1][]{#1}%
\maketitle
\begin{center}
    \centering
    \includegraphics[width=1\textwidth]{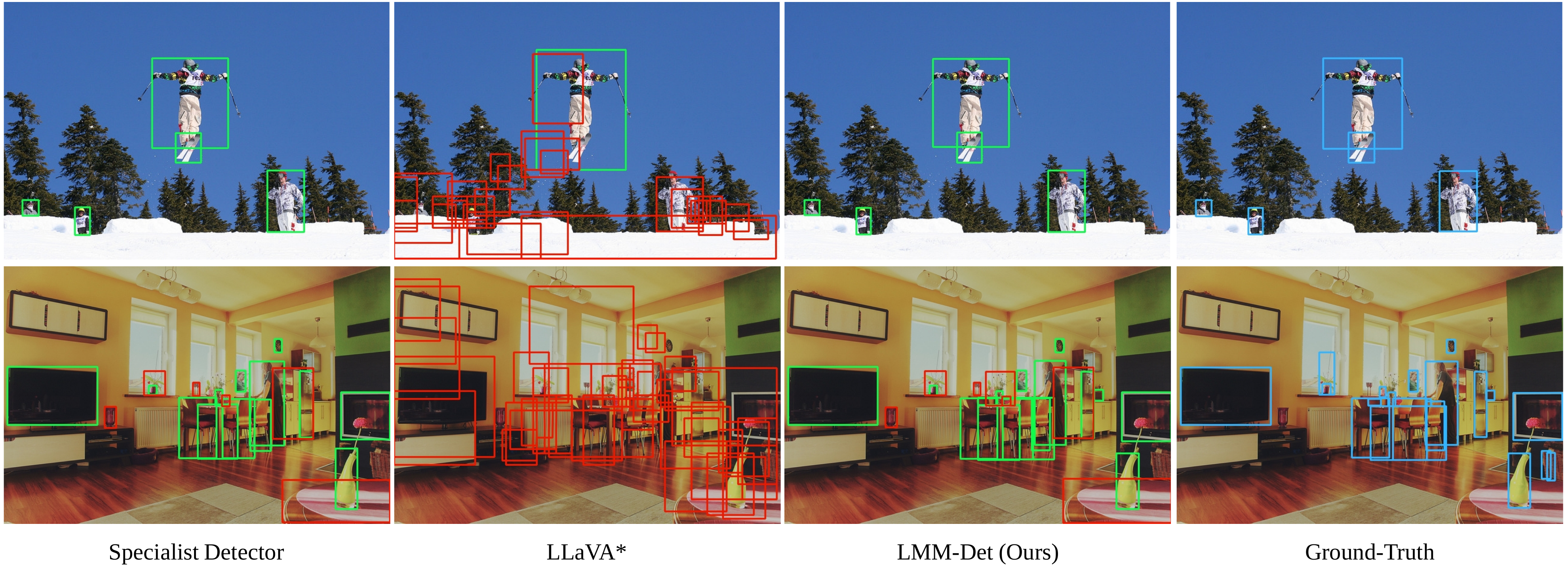}
    \captionof{figure}{
    Visualizations of a specialist detector and large multimodal models for object detection on the validation set of COCO. For clarity, the green bounding boxes indicate the correctly predicted boxes, the red boxes represent the prediction errors, and the blue boxes denote the ground truth labels. Specifically, we employ Salience-DETR~\cite{hou2024salience} as the specialist detector. Following the prompt style of RefCOCO, we query the trained LLaVA-7B~\cite{llava} with a sequence of questions, each targeting a single class category, which is called LLaVA$^*$. In addition, a predicted box is deemed correct if it meets the following criteria: (1) The intersection over union (IoU) between the predicted box and the ground truth (GT) box is greater than 0.5; (2) The predicted class label matches the GT class label.
    As illustrated in the figure above, our approach (LMM-Det) enables a large multimodal model to perform object detection, achieving comparable results with state-of-the-art specialist detection methods. It is noteworthy that LMM-Det exhibits outstanding object detection capabilities without relying on additional specialist modules within its architecture.
    The detailed predicted labels of all images can be referred to the Figure~\ref{appendix_detail_pred_label} in Appendix.
    }
    \label{fig:first_fig_llava_vs_llava_detection_true_false}
\end{center}%
}]

\begin{abstract}
\blfootnote{\textsuperscript{*} Equal Contribution}
\blfootnote{\textsuperscript{$\dag$} Corresponding Author, E-mail: lengdawei@360.cn}

Large multimodal models (LMMs) have garnered wide-spread attention and interest within the artificial intelligence research and industrial communities, owing to their remarkable capability in multimodal understanding, reasoning, and in-context learning, among others.
While LMMs have demonstrated promising results in tackling multimodal tasks like image captioning, visual question answering, and visual grounding, the object detection capabilities of LMMs exhibit a significant gap compared to specialist detectors.
To bridge the gap, we depart from the conventional methods of integrating heavy detectors with LMMs and propose LMM-Det, a simple yet effective approach that leverages a \textbf{L}arge \textbf{M}ultimodal \textbf{M}odel for vanilla object \textbf{Det}ection without relying on specialized detection modules.
Specifically, we conduct a comprehensive exploratory analysis when a large multimodal model meets with object detection, revealing that the recall rate degrades significantly compared with specialist detection models. 
To mitigate this, we propose to increase the recall rate by introducing data distribution adjustment and inference optimization tailored for object detection. We re-organize the instruction conversations to enhance the object detection capabilities of large multimodal models.
\cam{We claim that a large multimodal model possesses detection capability without any extra detection modules. Extensive experiments support our claim and show the effectiveness of the versatile LMM-Det.}
\cam{The datasets, models, and
codes are available at https://github.com/360CVGroup/LMM-Det.}
\vspace{-1.5em}
\end{abstract}
\section{Introduction}
\label{sec:intro}

Large multimodal models (LMMs) \cite{llava,flamingo,idefics3,yao2024minicpm,internvl2_5,Qwen2-VL,damonlpsg2023videollama} have recently attracted substantial attention due to their exceptional capabilities in comprehending and processing multiple forms of data, such as texts, images, speech, and videos. LMMs have shown notable effectiveness in bridging the gap between different modalities, \xie{enabling more efficient human-machine interaction.}
In particular, LMMs have exhibited strong performance in various multimodal tasks like image captioning \cite{li2023blip}, visual question answering (VQA) \cite{zhu2023minigpt}, and visual grounding \cite{chen2023shikra}. 
In the case of image captioning, a GPT assistant generates detailed textual descriptions for user-input images.
Moreover, the goal of VQA and visual grounding is to produce detailed \xie{and contextually appropriate responses} that meet the user's requirements.
These tasks demonstrate that large multimodal models are capable of aligning images and text at a fine-grained level, which holds \xie{significant} potential in fine-grained image-text understanding tasks like object detection and segmentation.

\xie{Generally, object detection is a fundamental task in computer vision, serving as an extension of classification and a foundation for segmentation.}
Furthermore, it is essential for effective visual understanding to acquire the location and category of objects.
\xie{Despite the impressive performance of large multimodal models in various multimodal tasks, their object detection capabilities remain underexplored in recent research. Compared to traditional state-of-the-art detection methods,} there is a significant gap in LMMs.

\xie{To bridge this gap, existing methods \cite{wu2024visionllm2,ma2024groma} attempt to integrate additional modules into LMMs, such as a specialized detection model \cite{liu2023groundingdino,ren2024grounding} or a region proposal network (RPN) \cite{fasterrcnn}. While these approaches can achieve detection capabilities in user dialogues, they are limited by the performance of the extra modules and introduce additional latency during inference.}
More importantly, they do not \xie{fully explore} the potential of LMMs to perform object detection tasks \xie{independently}.
\xie{On the other hand, some studies \cite{peng2023kosmos,llavagrounding,chen2023shikra,GenerateU,hanoona2023GLaMM,you2023ferret} make LMMs directly output object categories and bounding box coordinates, demonstrating detection abilities in visual grounding tasks such as the referring expression comprehension (REC) and phase grounding.} For instance, KOSMOS-2~\cite{peng2023kosmos} integrates grounding capabilities into downstream applications and provides the corresponding bounding boxes.
Shikra~\cite{chen2023shikra} designs a simple architecture without the need for pre-/post-detection modules and external plug-in models to unlock LMMs' grounding abilities.
\xie{However, it is non-trivial for these works to finish vanilla object detection, which requires localizing and classifying all objects within an image.}

\xie{The observation that large multimodal models exhibit fine-grained image-text alignment capability and possess preliminary detection capabilities (\eg REC), motivates} us to further investigate this phenomenon. Our goal is to explore the capability of LMMs in performing object detection tasks without relying on additional dedicated detection modules.
\xie{To this end, we highlight the following inherent challenges when large multimodal models meet with object detection.}

Most existing large multimodal models~\cite{chen2023shikra,peng2023kosmos,ma2024groma} evaluate their detection capabilities on RefCOCO~\cite{refcoco} instead of specialist detection benchmarks like COCO~\cite{coco}, 
\xie{despite using large-scale object detection datasets (\eg Object365~\cite{shao2019objects365}, LVIS~\cite{gupta2019lvis}, OpenImages~\cite{openimages} or COCO) for pre-training. Moreover, customized instruction data for \ljc{vanilla} object detection is often ignored during the instruction tuning stage, leading to poor localization and identification of objects within images. This hinders practical applications in fields such as robotics, autonomous driving, and augmented reality \cite{ma2024groma}.}

\xie{On the other hand, large multimodal models typically generate a small number of proposal bounding boxes, resulting in poor object detection performance. For instance, LLaVA~\cite{llava} shows suboptimal performance in detecting all objects} as illustrated in Figure~\ref{fig:first_fig_llava_vs_llava_detection_true_false}, using the following prompt: \cam{``Provide the bounding box coordinate of the region this sentence describes if region exists in the image: {<category>}''}.
\xie{For each query, LLaVA generates only a few bounding boxes, and most of them are inaccurate, resulting in a low recall rate for the object detection task.}

In this paper, a comprehensive analysis is conducted to investigate the performance gap between specialist object detection models and LMMs.
Specifically, we evaluate the large multimodal models on COCO, a specialist object detection benchmark. Through sample visualization and distribution comparison, we claim that the root cause of underperformance in the object \cam{detection} task is the low recall rate.
To address this, we present LMM-Det, a simple \xie{yet} effective approach to make large multimodal models excel in object detection.
\ljc{We propose to increase the recall rate by adjusting the training data distribution since we observe that the trained model tends to approximate this distribution. Moreover, we introduce inference optimization and present an instruction-tuning dataset.}
To summarize, our main contributions are as follows:

\begin{itemize}[leftmargin=*]
\item We focus on exploring the detection ability of large multimodal models to \xie{unlock their full potential} in practical applications. To this end, we provide a comprehensive analysis to facilitate the detection performance improvement for large multimodal models.
\item We propose a simple \xie{yet effective} approach named LMM-Det without any extra detection modules. \ljc{We introduce data distribution adjustment and inference optimization to increase the inherent recall rate when large multimodal models meet with object detection.}
\item \ljc{Extensive experiments demonstrate that LMM-Det not only exhibits detection capability but also preserves inherent multimodal capabilities like captioning and VQA.}
\end{itemize}

\section{Examination of LMMs in Object Detection}
\label{sec_examination}
To evaluate and analyze the detection performance of large multimodal models (LMMs), we first design a series of experiments from the perspective of data scale and image resolution. We then provide an in-depth analysis to facilitate better adaptation of LMMs to object detection. In this section, we use the standard large multimodal model LLaVA-7B \cite{llava} for all experiments. We choose RT-DETR~\cite{lv2023detrs} as the representative specialist detector for comparison.

\begin{figure}[t]
    \centering
	\includegraphics[width=0.48\textwidth]{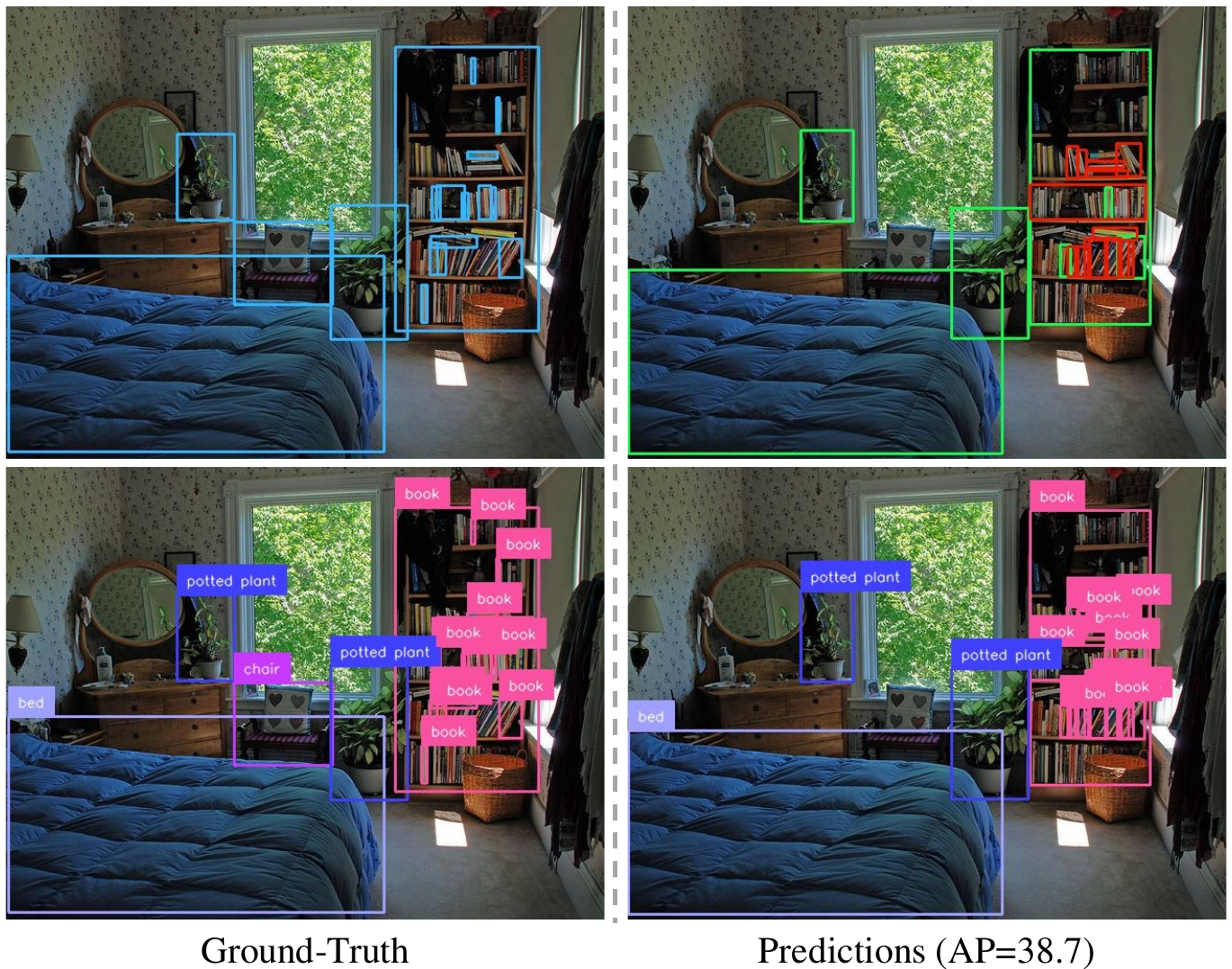}
	\caption{Visualizations of ground-truth and predicted bounding boxes, generated by the model in the 5th row of Table~\ref{tab:preliminary_lmm}.
	}
	\label{fig:examination_vis}
\end{figure}

\subsection{Exploratory Experiments}
\label{sec:Exploratory Experiments}
This section conducts exploratory experiments to \xie{assess the detection capabilities of LMMs} and the implementation details are provided in Section~\ref{appendix_stat_coco} of Appendix.

\textbf{Zero-shot on COCO.} A straightforward way to examine the detection performance of LMMs is through zero-shot evaluation. Thus, we first perform zero-shot experiments on COCO using LLaVA~\cite{llava}.
\xie{The qualitative results are shown in Figure~\ref{fig:first_fig_llava_vs_llava_detection_true_false}, while the quantitative results are provided in the first row of Table~\ref{tab:preliminary_lmm}.}
The experimental results reveal that LLaVA exhibits poor detection performance on the COCO dataset in a zero-shot setting, \xie{likely due to the absence of COCO data} during its training.

\textbf{\xie{Supervised Fine-tuning with COCO.}} 
To verify the above claim, we then leverage COCO for supervised fine-tuning.
As illustrated in the second and third rows of Table~\ref{tab:preliminary_lmm}, the incorporation of detection data (\ie COCO) yields a notable enhancement in AP, although there remains scope for further improvement. Inspired by the fact that expanding the data scale in conventional object detection methods is an effective technique to improve detection accuracy, we attempt to augment with additional detection data to make large multimodal models excel in object detection.

\textbf{Expand Detection Data.}
As aforementioned, we expand the detection data by incorporating Object365 during training.
From the fourth row of Table~\ref{tab:preliminary_lmm}, the addition of more data fails to provide a substantial boost in performance. The possible reasons are: (1) the low input image resolution is insufficient to perform object detection; (2) the instruction organization of COCO data needs to be improved.

\textbf{Upsample Image Resolution.} To \xie{ensure} a fair comparison with the specialist detector, we employ interpolation to increase the image resolution from 336 to 644, thus aligning it with the detector's native resolution.
\cam{As shown in rows 5-6 of Table~\ref{tab:preliminary_lmm}, the integration of Object365 using a higher input resolution of 644×644 indeed leads to a noticeable improvement in detection performance.
}
\cam{However, LLaVA's performance remains significantly inferior to that of the traditional specialist model, despite training on a similar resolution and an identical scale of detection data.}

\begin{figure}[t]
    \centering
	\includegraphics[width=0.49\textwidth]{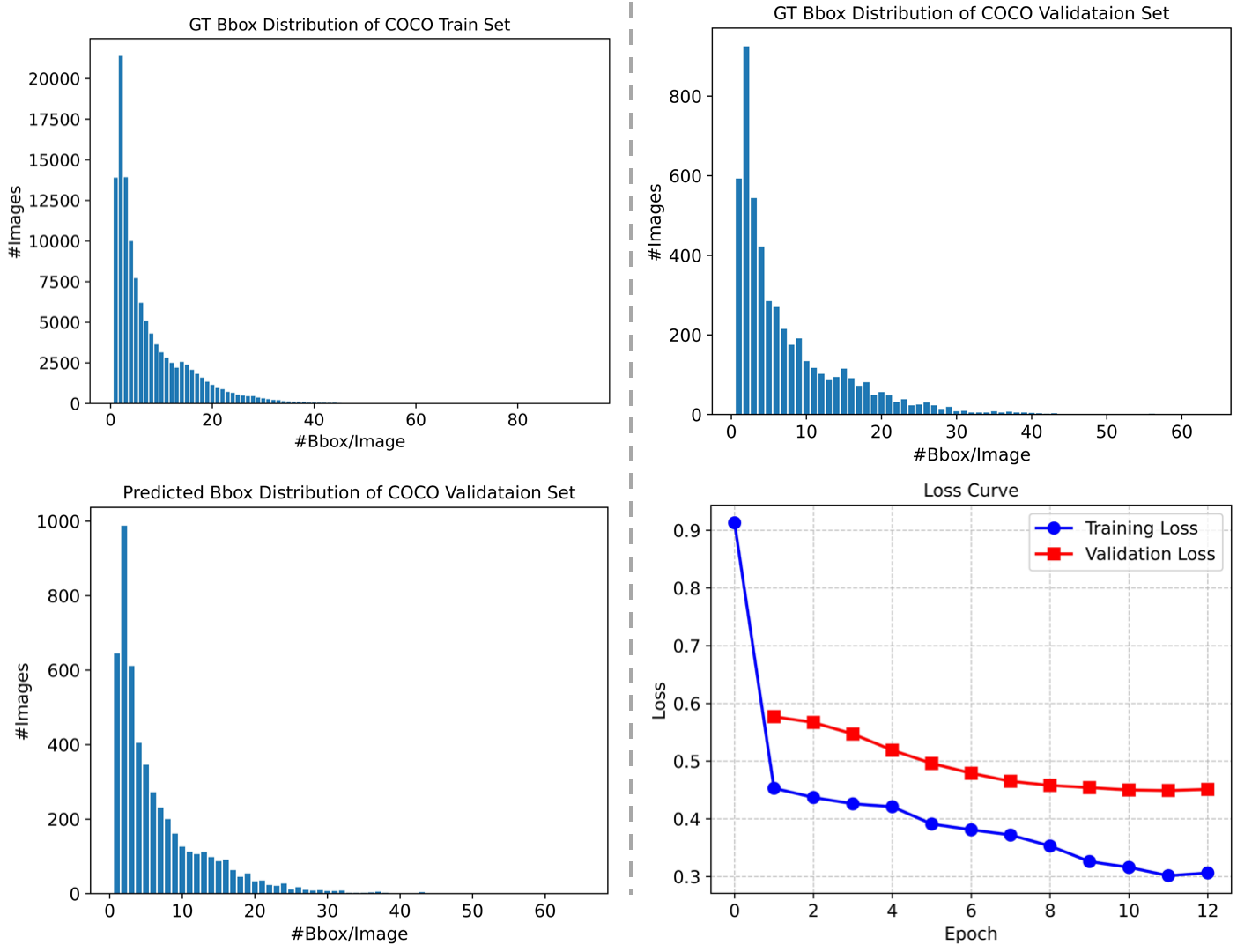}
	\caption{Bounding box (Bbox) distribution of ground-truth and predictions. We employ the model (AP=38.7) in the 5th row of Table~\ref{tab:preliminary_lmm} to generate boxes. In addition, we provide the loss curve to confirm that the model is not overfitting.
	}
	\label{fig:examination_distribution}
\end{figure}

\begin{table}[t]
  \centering
  
  \resizebox{0.48\textwidth}{!}{
    \begin{tabular}{lcccc|ccccc}
    \toprule
    Method & SFT data & COCO & Object365 & Res. & AP & AP$_{50}$ & AP$_{75}$ & AR${@100}$  \\
    \midrule
    \multirow{6}*{LLaVA} & $\yes$ & $\no$ & $\no$  & 336 & 0.2 & 0.6 & 0.2 & 11.2 \\
    & $\yes$ & $\yes$ & $\no$ & 336 & 14.5 & 29.5 & 12.7 & 24.3 \\
    & $\no$ & $\yes$ & $\no$ & 336 & 16.0 & 31.2  & 14.4 & 26.7 \\
    & $\no$ & $\yes$ & $\yes$ & 336 & 15.6 & 28.5 & 15.1 & 21.5 \\
    & $\no$ & $\yes$ & $\no$ & \cam{644} & \cam{17.7} & \cam{30.3} & \cam{17.5} & \cam{26.9} \\
    & $\no$ & $\yes$ & $\yes$ & 644 & 38.7 & 55.8 & 41.3 & 50.5 \\
    \midrule
    \lightgray{Specialist Detector} & $\no$ & $\yes$ & $\yes$ & \lightgray{640} & \lightgray{49.2} & \lightgray{66.6} & \lightgray{53.5} & \lightgray{70.9} \\
    \bottomrule
    \end{tabular}
    }
    \caption{Examinations of LLaVA-7B on the validation set of COCO. ``SFT data'' denotes the 665K instruction data during supervised fine-tuning (SFT) in LLaVA. ``Res.'' is image resolution.}
    \label{tab:preliminary_lmm}
\end{table}

\subsection{Phenomenon Analysis}
\xie{To narrow the performance gap between LMMs and specialist detection models, we investigate the underlying causes.}

\textbf{Visualizations of Predictions and GT.} \ljc{We visualize the generated bounding boxes with the corresponding categories and the ground-truth (GT) in the bottom row of Figure~\ref{fig:examination_vis}.}
The visualization results are surprisingly satisfactory despite being 10.5\% lower than the specialist model on AP.
\ljc{According to the criteria specified in Figure~\ref{fig:first_fig_llava_vs_llava_detection_true_false}, 
we plot boxes of different colors and provide the results in the top row of  Figure~\ref{fig:examination_vis}.
}
We observe that the red bounding boxes, deemed as false positives, are actually with correct predicted labels. Meanwhile, the labels for ground truth (GT) are not fully annotated, \eg ``book''.
Furthermore, our statistical results (as shown in Table~\ref{appendix_stat_coco} in Appendix) indicate that the average number of bounding boxes per image in \xie{both the training and validation sets of COCO is approximately 7}, which also matches the average number of boxes produced by the trained model. One possible reason is that the trained model has \xie{adapted to} the data distribution of the COCO dataset.

\textbf{Distribution \cam{Comparison}.} We provide the box distribution for predictions and ground truth to further analyze the above observations. From Figure~\ref{fig:examination_distribution}, \xie{the predicted box distribution approximates the training set of COCO. However, incomplete ground-truth annotations lead to premature truncation of predictions, resulting in a small number of generated bounding boxes. }
In addition, the current simple auto-regressive training recipe results in predicting fewer bounding boxes than specialist detection models.

Actually, traditional object detection methods maintain an appropriate recall rate on proposals (candidate regions), such as 300 proposals in Faster RCNN~\cite{fasterrcnn} or 900 proposals in H-Deformable-DETR~\cite{jia2022detrs} to balance detection performance and computational costs.
In this case, an insufficient recall rate can substantially degrade the detection performance.
However, it is challenging for a large multimodal model to function as a Region Proposal Network (RPN)~\cite{fasterrcnn} and generate a large number of high-quality proposals due to the inherent limitations of LMMs' next-token prediction loss under incomplete GT annotations. 
\xie{Therefore, increasing the recall rate is crucial for improving the overall detection performance of large multimodal models.}


\section{LMM-Det}
\ljc{This section introduces LMM-Det, which enhances LMMs' detection capabilities by increasing the recall rate. We first introduce the model architecture in Section~\ref{model_arch}. The advancements of LMM-Det consist of data distribution adjustment (Section~\ref{data_distribution_adjust}) and inference optimization (Section~\ref{infer_optim}).
}

\begin{figure}[t]
    \centering
	\includegraphics[width=0.48\textwidth]{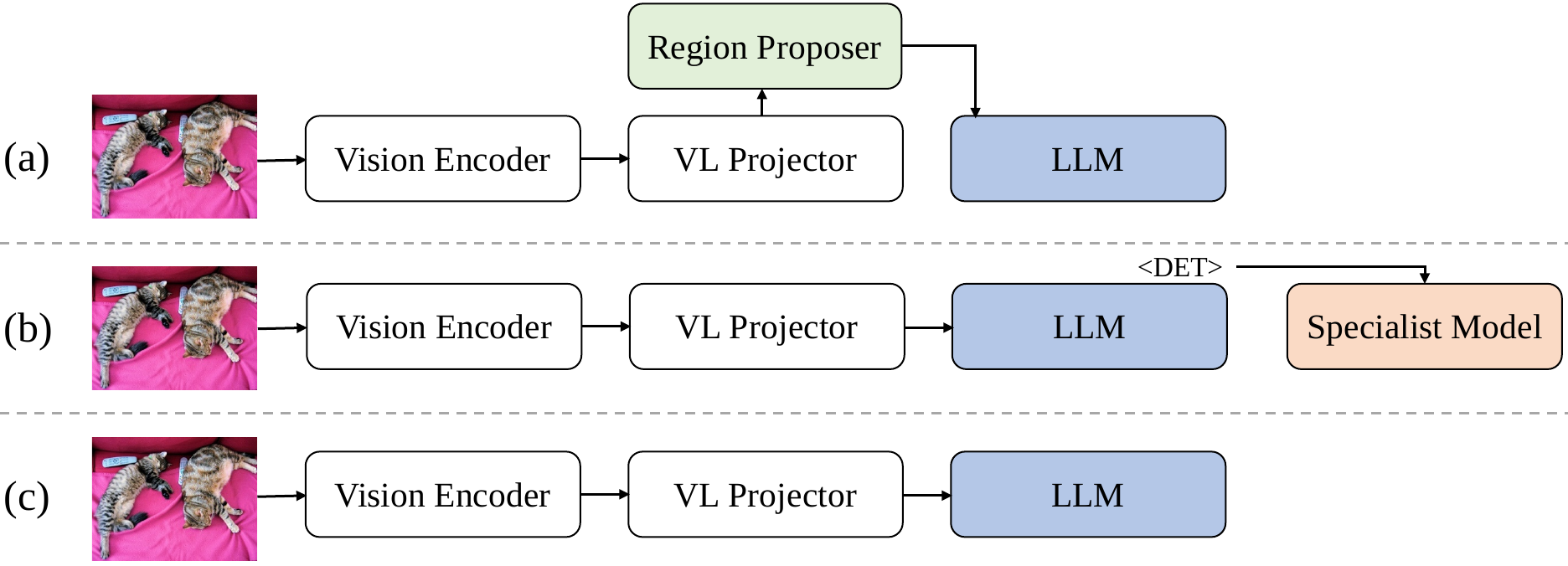}
	\caption{Overview of the proposed LMM-Det. Compared to the other LMMs with extra modules like region proposal generators (a) or specialist detection models (b), LMM-Det (c) enables LMMs to unlock the detection capability in a straightforward manner.}
	\label{fig:overview}
\end{figure}

\subsection{Preliminary: Model Architecture}
\label{model_arch}
\ljc{As illustrated in Figure~\ref{fig:overview} (c), LMM-Det consists of a visual encoder, a projector, and a large language model.}

\textbf{Visual Encoder.} LMM-Det employs the OWLv2-ViT model~\cite{minderer2024scaling} as the visual encoder, which supports high-resolution image input (1008 × 1008) and captures detailed information for object detection.
We do not compress the sequence of visual tokens and feed them \xie{directly} into LMMs.

\textbf{Projector.} 
\xie{
LLaVA~\cite{llava} demonstrates that a linear projector is sufficient for vision language alignment. Additionally, Honeybee~\cite{cha2024honeybee} shows that a linear projector can retain all the local context of visual features through a one-to-one projection without loss. Therefore, we use the linear projector to map visual features into the embedding space of text tokens.}

\textbf{Large Language Model.} We \xie{utilize} Vicuna-1.5-7B with a max sequence length of 16,000 as the large language model.
LMM-Det is trained to perform token prediction using a paradigm of language modeling loss. Formally, given an image and the well-organized instruction text tokens, we maximize the following optimization problem:

\begin{align}
    \max_{\theta} \sum_{i=1}^{L} \log p_{\theta}(\widetilde{\by}_i|\bx_{v},\bx_{t},\by_{1:i-1}),
\end{align}

where $\theta$ is the trainable parameters, $\bx_{v}$ is the visual tokens from the visual encoder and the projector, $\bx_{t}$ is the instruction text tokens from the tokenizer of the large language model, $\by_{1:i-1}$ is the answer tokens in the instruction data before the current prediction token $\widetilde{\by}_i$, and $L$ is the length of the output sequence.

\subsection{\ljc{Data Distribution Adjustment}}
\label{data_distribution_adjust}
\ljc{
As aforementioned in Section~\ref{sec_examination}, the incomplete ground-truth annotations in the training set may lead to premature truncation of predictions in large multimodal models. To mitigate this issue, we adjust the data distribution of the training set since the trained large multimodal model tends to approximate this distribution. Moreover, the process of data distribution adjustment should achieve the goal of increasing the recall rate to enhance the detection performance of large multimodal models.
}
\ljc{
To this end, a possible way is to use a self-training strategy, \ie generating pseudo labels by large multimodal models and training them with ground-truth labels iteratively. However, the generation of high-quality pseudo labels may face challenges in large multimodal models that do not possess strong object detection abilities. In this work, we use a semi-supervised learning strategy to adjust the training data distribution, increase the recall rate, and thus improve LMMs' detection accuracy.
}

\ljc{
Specifically, the data distribution adjustment of LMM-Det comprises three steps: 1) Pseudo Label Generation. This step aims to prepare more high-quality data for incomplete ground-truth annotations in the training set. 2) Data Merging and Design. The crafted pseudo labels and original ground-truth labels are merged and then designed to apply object detection to LMMs. 3) Token Representation Verification. We investigate different token representations to represent coordinates and confidence scores.
}

\ljc{
\textbf{Pseudo Label Generation.} We augment the ground-truth labels with high-quality pseudo-labels, thereby enhancing the annotation diversity. Concretely, we craft several pseudo-labels using a pre-trained specialist detector (\ie Salience-DETR~\cite{hou2024salience}). In this sense, we develop substantial proposals to improve the overall detection performance of large multimodal models.
Notably, our modification only involves scaling the data, while the architecture of our LLM-Det does not rely on additional detection modules throughout both the pre-training and supervised fine-tuning stages.
}

\ljc{
\textbf{Data Merging and Design.} To remove the redundant boxes and further improve the data quality, 
we merge the generated pseudo-labels and the original ground-truth labels by performing Non-Maximum Suppression (NMS).
Moreover, the existing data organization is non-trivial to compute average precision on COCO due to the absence of the confidence score of each predicted box. In practice, we empirically observe that the confidence score computed by the average probability of each coordinate token (after softmax) is not distinguishable.
To address this, we make a large multimodal model output both the coordinates and the corresponding confidence score instead of only the coordinates. We assign a confidence score of 1 to the ground-truth labels, whereas the confidence scores of the pseudo-labels are determined by the pre-trained specialist detector.
}

\ljc{
\textbf{Token Representation Verification.} 
We examine how to represent the coordinates and their corresponding confidence scores. We explore two alternatives: directly outputting token predictions and expanding the vocabulary with extra tokens.
The first approach increases the length of the model's output tokens without requiring additional training of vocabulary embeddings, whereas the second approach achieves the opposite, necessitating extra training of vocabulary embeddings but resulting in shorter output tokens. Experiments in Section~\ref{sec:ablation study} show that the first approach achieves better detection accuracy. Thus, it is adopted for all experiments.
}

\ljc{
By adjusting the training data distribution, LMM-Det can generate more accurate and numerous bounding boxes, which can increase the recall rate and then improve the overall detection performance. In addition, we empirically show that LMM-Det can not only approximate the detection data distribution but also maintain the original capabilities to approximate other data distributions like image captioning and visual question answering, as shown in Section~\ref{sec:versatile LMM-Det}.
}



\subsection{Inference Optimization Tailored for Detection}
\label{infer_optim}
\ljc{
At the inference stage, a straightforward approach for detecting all objects within an input image in large multimodal models is to predict all bounding boxes simultaneously, such as InternVL 2.5~\cite{internvl2_5}. However, we empirically observe that this approach is non-trivial to generate enough proposals with a fixed sampling strategy of LMMs. One possible reason is that current LMMs are hard to process all fine-grained image details in a single prediction step.
We attempt to optimize this solution to make LMM-Det output more proposals to increase the recall rate at inference.
}

\ljc{Specifically, we compromise on computational cost to obtain better detection accuracy of LMM-Det. Instead of outputting all bounding boxes in a single step, we make LMM-Det output all objects that belong to only one category in a single step and repeat this step several times. That is, LMM-Det predicts the bounding boxes for each category independently. To maintain consistency between inference and training, we re-organize the instruction conversations to adopt a class-specific prediction strategy.
\cam{We illustrate the re-organized data in Figure~\ref{fig:prompt_instruct_dataset} of Appenddix.}
}

\begin{table*}[t]
      \centering
      \resizebox{1\textwidth}{!}{
        \begin{tabular}{lcccccccccccc}
        \toprule
        Method & \cam{Visual Backbone} & LLM & w. Specialist  & AP & AP$_{\text{50}}$ & AP$_{\text{75}}$ & AP$_{\text{S}}$ & AP$_{\text{M}}$ & AP$_{\text{L}}$ & AR${\text{@100}}$  \\
        \midrule
        LLaVA~\cite{llava} & \cam{CLIP-L} & Vicuna-7B & $\no$ & 0.2 & 0.6 & 0.2 & 0.0 & 0.1 & 0.7 & 11.2 \\
        Shikra~\cite{chen2023shikra} & \cam{CLIP-L} & Vicuna-7B & $\no$ & 0.4 & 0.8 & 0.4 & 0.0 & 0.4 & 1.0 & 18.7 \\
        KOSMOS-2~\cite{peng2023kosmos}& \cam{CLIP-L} & MAGNETO-24-layers & $\no$ & 7.6 & 13.7 & 7.3 & 0.8 & 6.7 & 15.8 & 18.2 \\
        \ljc{InternVL-2.5}~\cite{internvl2_5}& InternViT-300M & Internlm2.5-7B & $\no$ & 11.8 & 18.4 & 12.0 & 3.6 & 13.0 & 23.4 & 27.5 \\
        Groma~\cite{ma2024groma} & DINOv2 & Vicuna-7B  & $\yes$ & 12.8 & 17.0 & 13.8 & 3.3 & 12.8 & 24.2 & 22.5 \\
        \textbf{LMM-Det (Ours)} & \cam{OWLv2-L} & Vicuna-7B & $\no$ & 24.5 & 34.7 & 26.3 & 15.4 & 27.4 & 37.3 & 46.6 \\
        \bottomrule
        \end{tabular}
        }
\caption{Zero-shot results on COCO compared with state-of-the-art large multimodal models. We do not report the zero-shot results of VisionLLM v2~\cite{wu2024visionllm2} since they do not release the pre-trained checkpoints. ``w. Specialist'' represents ``with specialist detection model''.}
\label{zeroshot}
\end{table*}

\begin{table*}[t]
      \centering
      \resizebox{1\textwidth}{!}{
        \begin{tabular}{lcccccccccccc}
        \toprule
        Method & \cam{Visual Backbone} & LLM & w. Specialist  & AP & AP$_{\text{50}}$ & AP$_{\text{75}}$ & AP$_{\text{S}}$ & AP$_{\text{M}}$ & AP$_{\text{L}}$ & AR${\text{@100}}$  \\
        \midrule
        Faster R-CNN~\cite{fasterrcnn} & RN50-FPN & - & $\yes$ &   40.2 & 61.0 & 43.8 & 24.2 & 43.5 & 52.0 & 54.0 \\
        Cascade R-CNN~\cite{cai18cascadercnn} & RN101-FPN & - & $\yes$  & 42.7 & 61.6 & 46.6 & 23.8 & 46.2 & 57.4 & -
        \\
        Deformable-DETR~\cite{zhu2020deformable} & RN50 & - & $\yes$ &  43.7 & 63.0 & 47.6 & 26.7 & 47.0 & 58.0 & 63.2 \\
        RT-DETR~\cite{lv2023detrs} & RN50 & - & $\yes$ &  55.3 & 73.4 & 60.0 & 38.0 & 59.9 & 71.6 & 74.4  \\
        H-Deformable-DETR~\cite{jia2022detrs} & Swin-L & - & $\yes$ &  56.1 & 75.1 & 61.3 & 39.2 & 60.4 & 72.4 & 73.1 \\
        Salience-DETR~\cite{hou2024salience} & FocalNet-L & - & $\yes$ &  57.3 & 75.5 & 62.4 & 40.9 & 61.8 & 74.5 & 75.4 \\
        \midrule
        Groma~\cite{ma2024groma} &  \cam{CLIP-L} & Vicuna-7B & $\yes$ &  32.4 & - & - & - & - & - & -  \\
        Groma~\cite{ma2024groma} &  DINOv2 & Vicuna-7B & $\yes$ &  43.6 & - & - & - & - & - & -  \\
        VisionLLM v2~\cite{wu2024visionllm2} & Swin-T & Vicuna-7B & $\yes$ &  56.3 $\downarrow$  & 74.3 & 61.6 & - &  - & - & - \\
        \lightgray{Grounding DINO~\cite{liu2023groundingdino}} & \lightgray{Swin-T} & \lightgray{-} & \lightgray{$\yes$}  & \lightgray{57.2} & \lightgray{-} & \lightgray{-} & \lightgray{-} \lightgray{-} & \lightgray{-} & \lightgray{-}  \\
        \cam{Griffon-13B~\cite{zhan2025griffonv1}} & \cam{CLIP-L} & LLaMA2-13B & $\no$ & 24.8 & 40.6 & 25.1 & 5.9 & 25.5 & 48.7 & - \\
        \cam{Griffon v2~\cite{zhan2024griffonv2}} & \cam{EVA2-CLIP-L} & LLaMA2-13B & $\no$ & 38.5 & 54.3 & 41.2 & 19.4 & 43.2 & 57.6 & - \\
        LLaVA*~\cite{llava} & \cam{CLIP-L} & Vicuna-7B & $\no$ & 38.7 & 55.8 & 41.3 & 20.1 & 43.6 & 57.3 & 50.5 \\
        \textbf{LMM-Det (Ours)} & \cam{OWLv2-L} & Vicuna-7B & $\no$ & 47.5 & 66.5 & 51.1 & 34.7 & 51.8 & 60.3 & 63.6 \\
        \textbf{LMM-Det\textsuperscript{$\dag$} (Ours)} & \cam{OWLv2-L} & Vicuna-7B & $\no$ & 47.1 & 66.2 & 50.5 & 35.0 & 51.6 & 60.1 & 63.1 \\
        \bottomrule
        \end{tabular}
        }
\caption{\cam{Fine-tuned results on COCO compared with traditional state-of-the-art detection models and large multimodal models relying on extra specialist detection models (w. Specialist).
VisionLLM v2 employs Grounding DINO as the additional specialist module for object detection, yet this integration compromises the original detection performance of Grounding DINO. LLaVA* denotes that we retrain LLaVA with Object365 and COCO. LMM-Det\textsuperscript{$\dag$} means that we apply the optional Stage IV to obtain a versatile LMM-Det.
}}
\label{finetune}
\end{table*}

\section{Experiments}
\subsection{Implementation Details}
\label{train_recipe}
\textbf{Training Recipe.} We train LMM-Det in three consecutive stages. Table~\ref{appendix_hyperparameter} in Appendix summarizes the hyperparameters of all stages.
\cam{We use 595K image-text pairs and 1.86M images to train LMM-Det in total.}
Training LMM-Det takes 176 hours on a cluster of 6 nodes, each equipped with 8 Nvidia H800 GPUs.
Below are the details of each stage.

\textbf{Stage I.}  
\xie{We align the vision and language modules by pre-training the projector while freezing the visual encoder and the large language model.} Specifically, we leverage the 595K image-text pairs employed in LLaVA~\cite{llava}.

\textbf{Stage II.}
We utilize a large-scale object detection dataset (\xie{\ie Object365~\cite{shao2019objects365}}) to pre-train the proposed LMM-Det. \xie{In this stage, we train the projector and the large language model while freezing the visual encoder.} 

\textbf{Stage III.} We re-organize an object detection instruction dataset (as shown in Figure~\ref{fig:prompt_instruct_dataset} of Appendix) built upon the COCO dataset, a widely-used benchmark for the object detection task. Then we fine-tune the projector and the large language model using this instruction data to further improve the detection capability of LMM-Det. 


\cam{\textbf{(Optional) Stage IV.} In this stage, the projector and large language model are trained with the visual encoder frozen, using the 665k LLaVA~\cite{llava} dataset alongside our proposed re-organized instruction data and adopting the same hyperparameters as Stage III. We denote it as LMM-Det\textsuperscript{$\dag$}.}

\textbf{Inference and Evaluation.} We evaluate the detection capability of LMM-Det on the validation set of COCO. For each category of the validation set, we construct the corresponding prompt for questioning. We gather all the predicted outputs and regard them as the final proposals.
\xie{We evaluate the mean average precision (mAP) to obtain quantitative results.} For simplicity, we denote AP as mAP. We take AP, AP$_{\text{50}}$, AP$_{\text{75}}$, AP$_{\text{S}}$, AP$_{\text{M}}$, AP$_{\text{L}}$, and AR${\text{@100}}$ as metric.


\subsection{Zero-shot Experiments}
\label{sec:zero-shot-exp}
In this section, we compare LMM-Det with state-of-the-art LMMs in a zero-shot setting.
We adopt their \xie{official trained} models and test them in a zero-shot way. The implementation details are given in the \ljc{Appendix}.
LMM-Det only employs two stages as described in Section~\ref{train_recipe} and discards Stage III.

\begin{figure*}[t]
    \centering
	\includegraphics[width=1\textwidth]{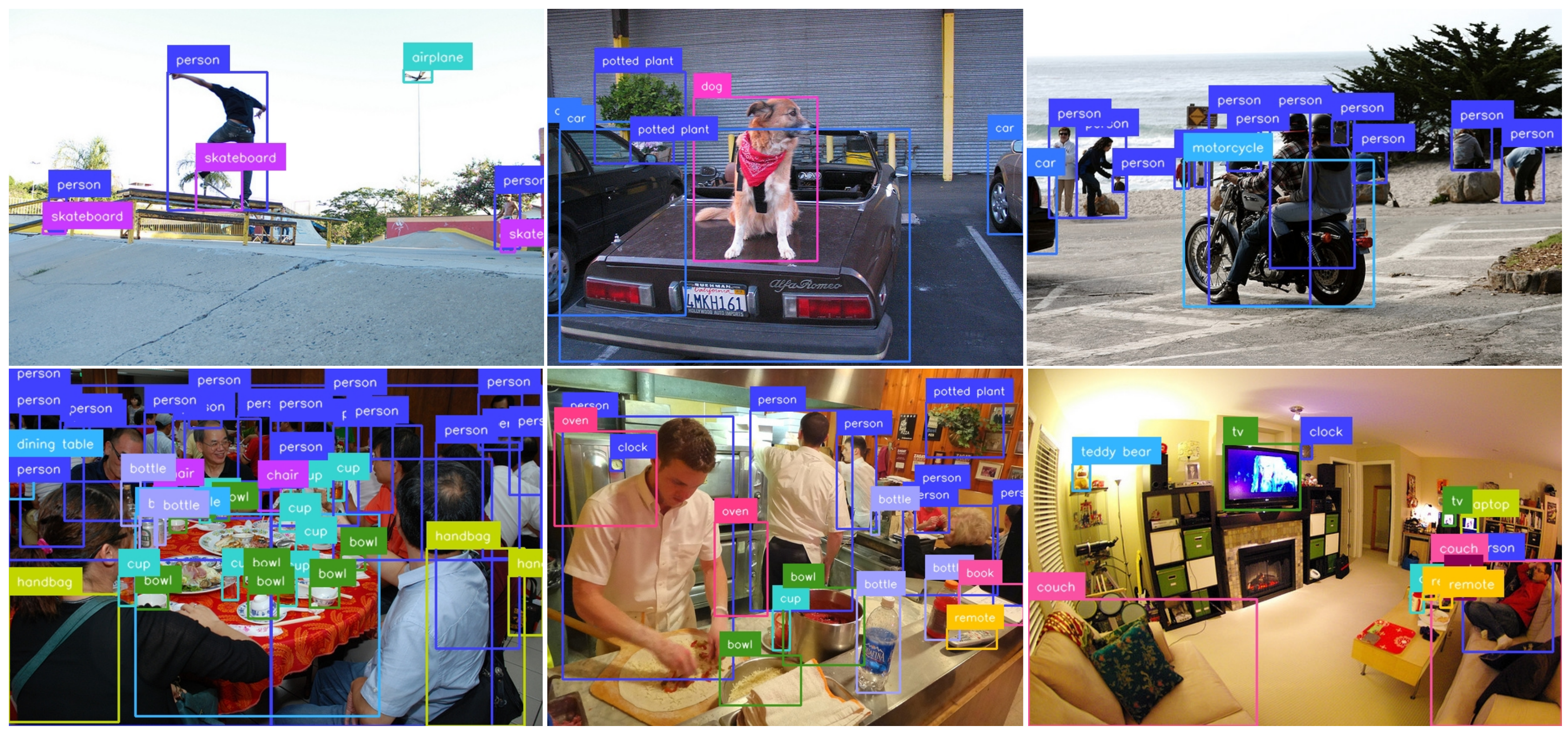}
	\caption{Visualizations of LMM-Det on COCO.
	}
	\label{vis_on_coco}
\end{figure*}

\xie{From Table~\ref{zeroshot}, LMM-Det achieves the best results on COCO in a zero-shot manner, demonstrating its potential for detection after training on detection data. LLaVA, without training on detection data, only achieves 0.2 AP on COCO. Models like \ljc{InternVL-2.5}, which utilizes a large amount of detection data, show better performance than those with limited data. Groma, which incorporates a specialist detection model, also performs well. Without extensive detection data and the specialist model, LMM-Det significantly outperforms other LMMs, validating our phenomenon analysis and the effectiveness of our proposed approach.}


\subsection{Fine-tuned Experiments}
We further fine-tune LMM-Det on COCO.
We compare LMM-Det with traditional detection models and LMMs relying on extra detection experts that can perform the object detection task (\ie Groma~\cite{ma2024groma} and VisionLLMv2~\cite{wu2024visionllm2}). 
\xie{Specifically, VisionLLMv2 uses Grounding DINO as the extra specialist detection model. For a comprehensive comparison, we also report the results of Grounding DINO.}

\xie{As shown in Table~\ref{finetune}, traditional state-of-the-art detection models outperform Groma, even when Groma is augmented with specialist detection models. Similarly, VisionLLMv2, despite incorporating Grounding DINO, exhibits a performance decline. We also retrain LLaVA using both COCO and Object365 datasets. The experimental results highlight a significant performance gap between traditional state-of-the-art detection models and large multimodal models. LMM-Det, however, narrows this gap, supporting the argument that LMMs can inherently perform object detection without the need for additional specialist detection models.}

\subsection{Versatile LMM-Det}
\label{sec:versatile LMM-Det}
\begin{figure*}[t]
    \centering
	\includegraphics[width=1\textwidth]{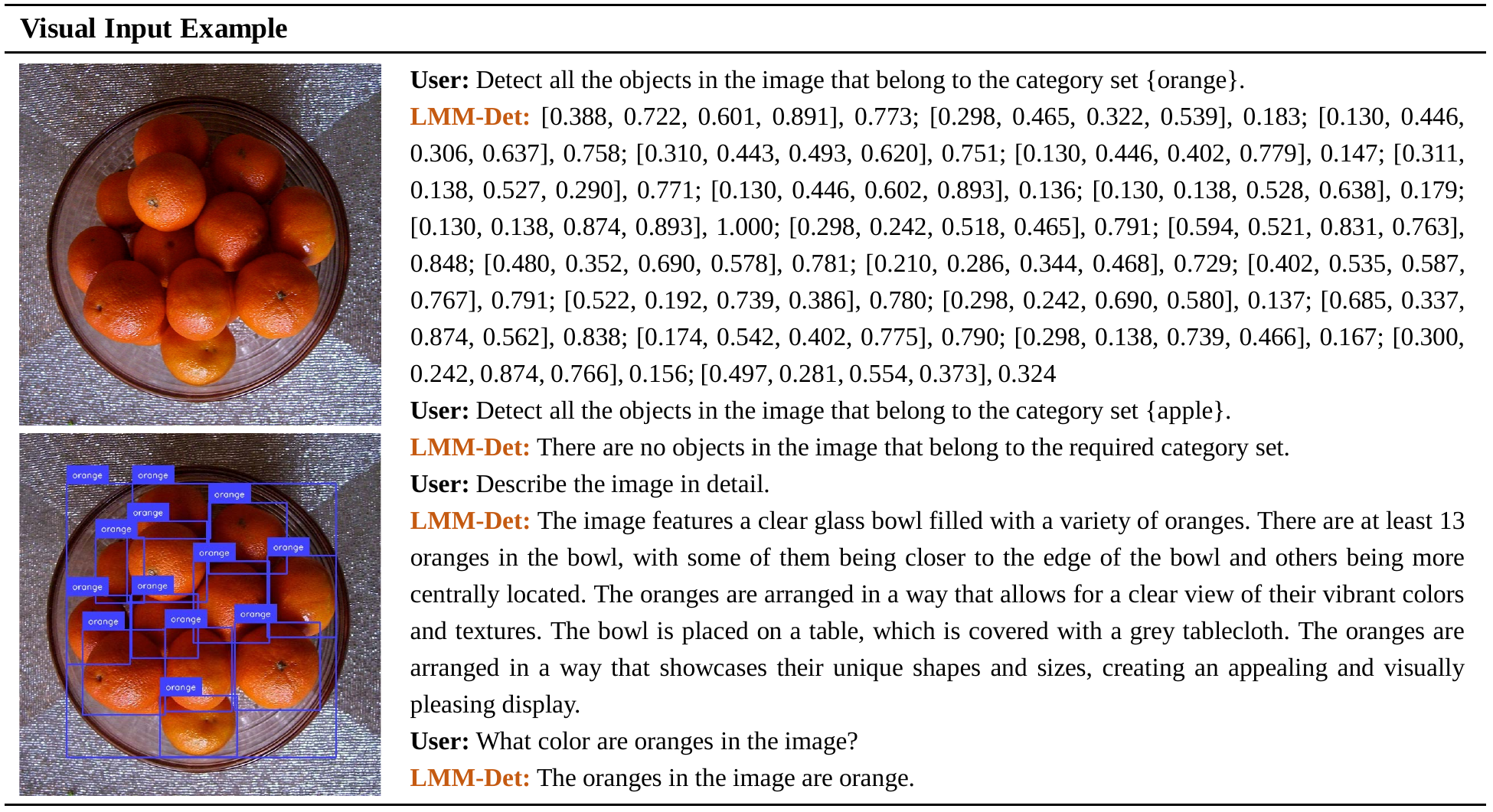}
	\caption{Chat example demonstrates that LMM-Det not only enables detection but also retains its intrinsic multimodal capabilities. The left-upper image represents the input, while the left-lower image illustrates the visualization of the output coordinates. For better visualization, we apply post-processing for the output coordinates by employing NMS with a threshold of 0.5.
	}
	\label{vis_on_cap_vqa}
\end{figure*}

\ljc{We provide visualizations of LMM-Det on the COCO validation set in Figure~\ref{vis_on_coco}, showcasing its detection performance without extra specialist detectors. Furthermore, the qualitative and quantitative results (as given in Figure~\ref{vis_on_cap_vqa} and Table~\ref{tab:quan_lmm_det}, respectively) demonstrate LMM-Det's versatility, supporting that \cam{LMM-Det\textsuperscript{$\dag$}} not only unlocks the detection capabilities but also maintains high performance in image captioning and VQA.} \cam{We provide more results in Table~\ref{appendix_more_griffon} in Appendix.}

\begin{table}[t]
\centering
\resizebox{1\linewidth}{!}{
\begin{tabular}{l|cccccc}
\hline
\multirow{2.5}{*}{Model} & COCO & \multicolumn{4}{c}{Image Captioning} & VQAv2 \\
\cmidrule{3-6} 
& AP & BLEU@4 & METEOR & CIDEr & SPICE & Accuracy \\
\hline
LLaVA & 0.2 & 29.4 & 29.3 & 108.9 & 23.6 & 78.5 \\
\cam{LMM-Det\textsuperscript{$\dag$}} & 47.1 & 26.7 & 28.1 &99.0 & 22.4 & 74.1 \\
\hline
\end{tabular}
}
\caption{Quantitative results for versatile \cam{LMM-Det\textsuperscript{$\dag$}}.}
\label{tab:quan_lmm_det}
\end{table}

\subsection{Ablation Study}
\label{sec:ablation study}


\textbf{Visual encoder.} We first replace the CLIP-ViT~\cite{CLIP} with OWLv2-ViT~\cite{minderer2024scaling} to further increase the input resolution in ablation study. As shown in the first row of Table~\ref{abation_study}, this strategy achieves a 3.4\% gain in AP. \cam{We further investigate the effectiveness of DINOv2~\cite{oquab2023dinov2} in Table~\ref{appendix_more_visual_encoder} of Appendix.}

\ljc{\textbf{Data Distribution Adjustment.} This approach aims to address the challenge of the insufficient recall rate.}
\xie{In Table~\ref{abation_study}, it effectively improves the recall rate and enhances overall detection performance .}
\ljc{We also draw the adjusted bounding box distribution in Figure~\ref{appendix_adjust_data_distribution} of Appendix.}

\begin{table}[t]
      \centering
      \resizebox{0.45\textwidth}{!}{
        \begin{tabular}{ccc|cccccc}
        \toprule
        \cam{OWLv2-ViT} & DDA & INO & AP & AP$_{\text{50}}$ & AP$_{\text{75}}$ & AR${\text{@100}}$  \\
        \midrule
        $\no$ & $\no$ & $\no$ & 38.7 & 55.8 & 41.3 & 50.5 \\
        $\yes$ & $\no$ & $\no$ & 42.1 & 57.8 & 45.8 & 51.3 \\
        $\yes$ & $\yes$ & $\no$ & 44.2 & 61.3 & 47.5 & 56.0 \\
        $\yes$ & $\yes$ & $\yes$ & 47.5 & 66.5 & 51.1 & 63.6 \\
        \bottomrule
        \end{tabular}
        }
\caption{Ablation Study. The baseline experiment in the first row is LLaVA*~\cite{llava}, which uses the CLIP-ViT as the visual encoder.
DDA: data distribution adjustment; INO: inference optimization.}
\label{abation_study}
\end{table}

\ljc{\textbf{Inference Optimization.} We further investigate the effectiveness of the inference optimization. From Table~\ref{abation_study}, this strategy results in a significant increase in AP from 44.2\% to 47.5\% and AR$\text{@100}$ from 56.0\% to 63.6\%. To better visualize the enhancement of recall rate, we extend Figure~\ref{fig:examination_vis} and show the qualitative results in Figure~\ref{appendix_after_enhancement} of Appendix.
}

\textbf{Effectiveness of extra vocabularies.} 
\ljc{We investigate different token representations to represent coordinates and confidence scores.}
\ljc{From Table~\ref{appendix_extra_vocab} in Appendix, directly outputting token achieves better performance than extra vocabularies,} which aligns with the findings of Shikra \cite{chen2023shikra}.


\textbf{Effectiveness of sampling strategy and inference cost.} 
\cam{We conduct ablation studies using greedy decoding, beam search, and top-p sampling during inference.}
As shown in Table~\ref{appendix_sampling_strategy} in Appendix, beam search with beam=2 achieves the best results but requires more inference time.
In particular, LMM-Det requires a computational time of approximately 4.0 seconds to process a single image.




\section{Related Work}
\label{related_work}

\subsection{Large Multimodal Models}

Pioneering works~\cite{li2023blip,xie2023ccmb,zhu2023minigpt,du2022glm,ye2023mplugowl,X-llm,lu2023ziya,chen2024allava,he2024bunny,tong2024cambrian1,xu2024llavacot,xie2025fg} about large multimodal models (LMMs) primarily focus on vision-language alignment and support fundamental multimodal tasks such as image captioning and visual question answering.
In this case, LMMs have demonstrated their considerable potential to perform multimodal tasks.

To further demonstrate the comprehensive capabilities of vision-language tasks, subsequent large multimodal models support resolution-sensitive tasks such as visual grounding and optical character recognition (OCR). These models leverage meticulously curated pre-training and instruction datasets, showcasing an effective strategy that highlights their robust capabilities.
This type of LMMs encompass a broad range of models~\cite{chen2023shikra,wei2023skywork,peng2023kosmos,hu2024docowl,wei2023vary,wei2024general,bai2023qwen,li2023monkey,zhang2024ferret,liu2024llavanext,hong2024cogvlm2,idefics3,yao2024minicpm,internvl2_5,wang2025iaa,wu2024deepseekvl2mixtureofexpertsvisionlanguagemodels}, such as Shikra, KOSMOS-2, Ferret v2, MiniCPM-V 2.6, DeepSeek-VL2, InternVL-2.5 and Qwen2.5-VL.
Compared to the coarse-grained visual question answering, the object localization capabilities of these LMMs are a crucial prerequisite for the effective deployment of LMMs in real-world applications such as agent, robotics, autonomous driving, and security surveillance. In this paper, we examine the detection potential of LMMs.

\subsection{Detection Capability Exploration for LMMs}
While LMMs show remarkable success on most vision-language benchmarks, they struggle to achieve comparable performance on object detection tasks.
\cam{Existing approaches~\cite{wu2024visionllm2, ma2024groma, GenerateU, internvl2_5, zhan2025griffonv1, zhan2024griffonv2} attempt to achieve object detection over large multimodal models.}
For example, VisionLLM-v2~\cite{wu2024visionllm2} introduces Grounding-DINO~\cite{liu2023groundingdino} into LMMs and uses a special token <DET> to perform object detection. Groma~\cite{ma2024groma} employs a region proposer to discover regions of interest to unlock the detection capabilities of LMMs.
Instead, this paper aims to investigate the object detection capability of large multimodal models without the help of specialist detection models or proposal networks.


\section{Conclusion and Limitations}
\label{conclusion}
LMM-Det has addressed the detection performance gap between LMMs and specialist detectors without relying on extra modules. 
\cam{Specifically, we provide a comprehensive exploration analysis and summarize that the key challenge is the insufficient recall rate. We enable LMM-Det to excel in object detection while preserving its inherent capabilities like caption generation and VQA.}
Besides, LMM-Det suffers from \xie{non-negligible inference time latency}, making it less competitive with traditional real-time detections.
In the future, we attempt to reduce the inference time of LMM-Det.

\section{\cam{Acknowledgements}}
\cam{We sincerely thank Shuaicheng Niu from Nanyang Technological University and Xiaole Zhu from 360 AI Research for their valuable discussions and feedback.}





{
    \small
    \bibliographystyle{ieeenat_fullname}
    \bibliography{reference}
}
\clearpage
\setcounter{page}{1}
\setcounter{equation}{0}
\setcounter{table}{0}
\setcounter{figure}{0}

\setcounter{section}{0}
\renewcommand\thesection{\Alph{section}}


\section{More Implementation Details.}
\label{sec:Appendix}

The statistical results of COCO are listed in Table~\ref{appendix_stat_coco}.

\begin{table}[h]
    \renewcommand\thetable{A}
      \centering
      \resizebox{0.36\textwidth}{!}{
        \begin{tabular}{l|ccc}
        \toprule
         & train & validation  \\
        \midrule
        \#images & 118,287 & 5,000 \\
        \#bboxes & 860,001 & 36,781 \\
        \#bboxes per image & 7.3 & 7.4 \\
        \bottomrule
        \end{tabular}
        }
\caption{Statistical results of COCO.}
\label{appendix_stat_coco}
\end{table}

The training hyperparameters of LMM-Det cross three stages are listed in Table~\ref{appendix_hyperparameter}.

\begin{table}[h]
    \renewcommand\thetable{B}
      \centering
      \resizebox{0.48\textwidth}{!}{
        \begin{tabular}{l|ccc}
        \toprule
        Configuration & Stage I & Stage II & Stage III  \\
        \midrule
        Training epochs & 1 & 5 & 12 \\
        Global batch size & 192 & 480 & 288 \\
        Learning rate & 1e-3 & 2e-5 & 2e-5 \\
        Learning rate schedule & \multicolumn{3}{c}{Cosine decay} \\
        Warmup ratio & 0.03 & 0.05 & 0.05 \\
        Weight decay & 0 & 0.05 & 0.05 \\
        Optimizer & \multicolumn{3}{c}{AdamW} \\
        Optimizer hyperparameters & \multicolumn{3}{c}{$\beta_1=0.9, \beta_2=0.999, \epsilon=1e-8$} \\
        Deepspeed ZeRO stage & ZeRO-2 & ZeRO-3 & ZeRO-3 \\
        Text max sequence length & 2k & 4k & 10k \\
        Training precision & \multicolumn{3}{c}{bf16} \\
        \bottomrule
        \end{tabular}
        }
\caption{Training hyperparameters in three stages.}
\label{appendix_hyperparameter}
\end{table}



\textbf{More details of the exploratory experiments.} \xie{In Section~\ref{sec:Exploratory Experiments},} we re-train LLaVA with detection data (\ie Object365 and COCO) using the same hyperparameters in Stage III of Table~\ref{appendix_hyperparameter} except for the text max sequence length \xie{, which is set to 2k in the exploratory experiments. During both training and inference stages, we output all prediction bounding boxes simultaneously. Additionally, we also predict bounding boxes for each category when conducting zero-shot detection on COCO. Unfortunately, this approach fails to improve detection performance and instead increases the number of incorrect prediction bounding boxes.} 

\textbf{More \ljc{implemetation details during training and inference}.} 
During training, we construct multi-turn conversations for each input image. To mitigate potential overfitting, we randomize each turn of the conversation and each bounding box in the target sequence for each training epoch. 
\cam{
For each image from either COCO or Object365, we construct both positive and negative conversations at a 1:1 ratio based on the number of existing categories in the image.
Specifically: 1) For an image containing $n$ visible categories (e.g., cat, dog), we generate $n$ positive instructions where the model is asked to output bounding boxes. 2) We then sample $n$ negative instructions by randomly selecting non-present categories from the remaining label set (\ie $80 - n$ for COCO, $365-n$ for Object365). 3) The maximum number of instruction rounds per image is capped at 80 for COCO and 365 for Object365, respectively. Importantly, we do not filter the Object365 dataset but retain all its categories and instances to preserve the model's broad detection capability.
}

\xie{During inference, we predict bounding boxes with confidence scores. Notably, the mean number of bounding boxes per image increases from 7 to 31 due to the integration of pseudo-labels and re-organized instruction data. However, the number of generated proposals is still lower than that of specialized models (e.g., 900 proposals in Salience-DETR) when calculating the AP. Therefore, we do not apply non-maximum suppression (NMS) and set a threshold for calculating AP and AR. For better visualization, we use a score threshold of 0.5 and NMS with a threshold of 0.5.}


\ljc{\textbf{More details of zero-shot experiments.} In Section~\ref{sec:zero-shot-exp}, we compare LMM-Det with a variety of large multimodal models in a zero-shot manner. We provide the detailed prompt for all models in Table~\ref{appendix_zero-shot-details}.
}
\ljc{
For simplicity, we omit the <image> token for KOSMOS-2 and Groma. We omit the special token in the prompt of QwenVL-2.5 such as <|im\_start|> and <|im\_end|>. In particular, we randomly select the templates of the referring expression comprehension (REC) task in Shikra to perform the zero-shot experiments on COCO. We illustrate an example for Shikra in Table~\ref{appendix_zero-shot-details}.
}

\textbf{More details of versatile LMM-Det.} \xie{In Section~\ref{sec:versatile LMM-Det}, the experiments demonstrate that LMM-Det not only unlocks object detection capabilities but also preserves inherent multimodal capabilities such as image captioning and visual question answering. Specifically, after training LMM-Det through the three stages outlined in the training recipe in Section~\ref{train_recipe}, we add a fourth stage. In this stage, we train the projector and large language model while freezing the visual encoder, using the 665K data from LLaVA~\cite{llava} and the proposed re-organized instruction data. We use the same hyperparameters as those in the fine-tuning stage of LLaVA.}


\section{More Quantitative Results.}
\label{appendix_more_qual_res}
\cam{This paper reveals the root cause for the poor performance of LMMs in the field of object detection (ODet). LMM-Det can process multimodal tasks (\eg ODet + referring expression comprehension (REC) + image captioning + visual question answering) while Griffon~\cite{zhan2025griffonv1} and Griffon v2~\cite{zhan2024griffonv2} focus on ODet+REC.
Table~\ref{appendix_more_griffon} further indicates that LMM-Det\textsuperscript{$\dag$} can also unify object detection with the REC task and shows mutual performance benefits (e.g., 81.4 $\Rightarrow$ 85.7).
}

\begin{table}[h]
\renewcommand\thetable{D}
      \centering
      \resizebox{0.4\textwidth}{!}{
        \begin{tabular}{c|ccc}
        \hline
        Model & COCO & RefCOCO val & MMStar \\
        \hline
        LLaVA-7B   &  0.2 & 81.4 & 30.3 \\ 
        Griffon-13B  &  24.8 & 88.0  & -\\ 
        Griffon v2-13B  &  38.5 & 89.6  & - \\ 
        LMM-Det\textsuperscript{$\dag$}-7B  & 47.1  & 85.7 & 32.1 \\ 
        \hline
      \end{tabular}
      }
\caption{More quantitative results for versatile LMM-Det\textsuperscript{$\dag$}.}
\label{appendix_more_griffon}
\end{table}

\begin{table*}[t]
    \renewcommand\thetable{C}
      \centering
      \resizebox{1\textwidth}{!}{
        \begin{tabular}{lcccccccc}
        \toprule
        Model & Multi-step & CLIP emb & Prompt  \\
        \midrule
        LLaVA~\cite{llava} & $\yes$ & $\no$ & <image>\textbackslash nProvide the bounding box coordinate of the region this sentence describes if region exists in the image: <category> \\
        Shikra~\cite{chen2023shikra} & $\yes$ & $\no$  & May I have the coordinates of <category> in <image>? \\
        KOSMOS-2~\cite{peng2023kosmos} & $\yes$ & $\no$ & <grounding> Where is the <category>? \\
        InternVL-2.5~\cite{internvl2_5} & $\no$ & $\yes$ & <image>\textbackslash nPlease detect and label all objects in the following image and mark their positions. \\
        Groma~\cite{ma2024groma} & $\no$ & $\yes$ & [grounding] Please summarize the content of this image in detail.\\
        LMM-Det (Ours) & $\yes$ & $\no$ & <image>\textbackslash nDetect all the objects in the image that belong to the category set <category>. \\
        \bottomrule
        \end{tabular}
        }
\caption{\ljc{Detailed prompt for performing zero-shot object detection task on COCO. ``Multi-step'' denotes whether to use multi-step inference to predict images. For each image, we construct 80 steps for LLaVA to predict the bounding boxes on COCO. ``CLIP emb'' represents whether to use CLIP embbedings. In this way, we map the unknown category to the pre-defined categories (\eg 80 categories on COCO).
}}
\label{appendix_zero-shot-details}
\end{table*}

\begin{figure*}[h]
    \renewcommand\thefigure{A}
    \centering
	\includegraphics[width=1\textwidth]{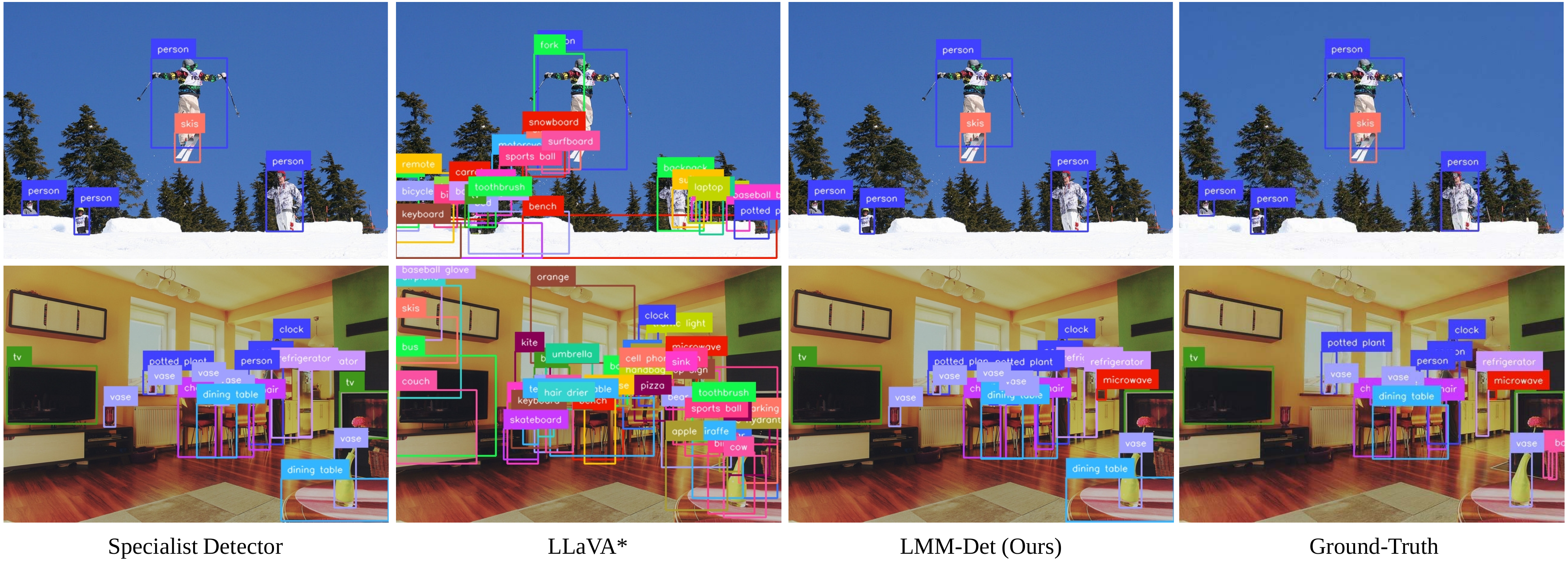}
	\caption{Detailed predicted labels of Figure~\ref{fig:first_fig_llava_vs_llava_detection_true_false} in the manuscript.
	}
	\label{appendix_detail_pred_label}
\end{figure*}

\cam{Table~\ref{appendix_more_visual_encoder} demonstrates that Owlv2-ViT works better than other visual encoders. In particular, the experiments (Tables (\ref{appendix_more_visual_encoder}, \ref{appendix_extra_vocab}, \ref{appendix_sampling_strategy}) are trained on only COCO for simplicity.
}

\begin{table}[h]
\renewcommand\thetable{E}
      \centering
      \resizebox{0.45\textwidth}{!}{
  \begin{tabular}{cccccc}
    \hline
    Model & Res. & mAP & AP$_{\text{50}}$ & AP$_{\text{75}}$ & AR${\text{@100}}$ \\
    \hline
    CLIP-ViT-L & 336  &  16.0 & 31.2 & 14.4 & 26.7 \\
    DINOv2-L & 224 & 12.3  & 24.5 & 11.3 & 19.1 \\ 
    OWLv2-ViT-L & 1008 &  32.6 & 50.5 & 34.4 & 43.1 \\ 
    \hline
  \end{tabular}
      }
\caption{More ablation studies of the visual encoder.
}
\label{appendix_more_visual_encoder}
\end{table}

The effect of extra vocabularies is shown in Table~\ref{appendix_extra_vocab}.

\begin{table}[h]
    \renewcommand\thetable{F}
      \centering
      \resizebox{0.4\textwidth}{!}{
        \begin{tabular}{lcccccccc}
        \toprule
        Techniques & AP & AP$_{\text{50}}$ & AP$_{\text{75}}$ & AR${\text{@100}}$  \\
        \midrule
        LMM-Det &  32.6 & 50.5 & 34.4 & 43.1 \\
        + extra vocabularies & 29.2 & 47.3 & 29.9 & 40.8 \\
        \bottomrule
        \end{tabular}
        }
\caption{The effectiveness of extra vocabularies.
}
\label{appendix_extra_vocab}
\end{table}

The effectiveness of the sampling strategy is listed in Table~\ref{appendix_sampling_strategy}.
\cam{We use greedy decoding as the inference sampling strategy for all experiments in the manuscript.}

\begin{table}[h]
    \renewcommand\thetable{G}
      \centering
      \resizebox{0.48\textwidth}{!}{
        \begin{tabular}{lcccccccc}
        \toprule
        Model & Techniques & AP & AP$_{\text{50}}$ & AP$_{\text{75}}$ & AR${\text{@100}}$ & Cost/Img \\
        \midrule
        \multirow{5}*{LMM-Det} & greedy decoding & 32.6 & 50.5 & 34.4 & 43.1 & 4.0 s \\
        & beam search (beam=2) & 33.0 & 51.3 & 34.9 & 43.7 & 8.2 s \\
        & beam search (beam=3) & 32.8 & 51.1 & 34.5 & 43.6 & 10.9 s \\
        & beam search (beam=4) & 24.6 & 38.2 & 25.8 & 32.5 & 13.8 s \\
        & top-p sampling & 27.6 & 45.6 & 28.0 & 41.7& 4.3 s \\
        \bottomrule
        \end{tabular}
        }
\caption{Inference Sampling strategy.
Cost/Img denotes the inference computational cost for one image, which is measured on one Nvidia H800 GPU without TensorRT. We average the inference cost of all images on COCO val set to obtain Cost/Img. 
}
\label{appendix_sampling_strategy}
\end{table}

\begin{figure*}[h]
\renewcommand\thefigure{B}
    \centering
	\includegraphics[width=1.0\textwidth]{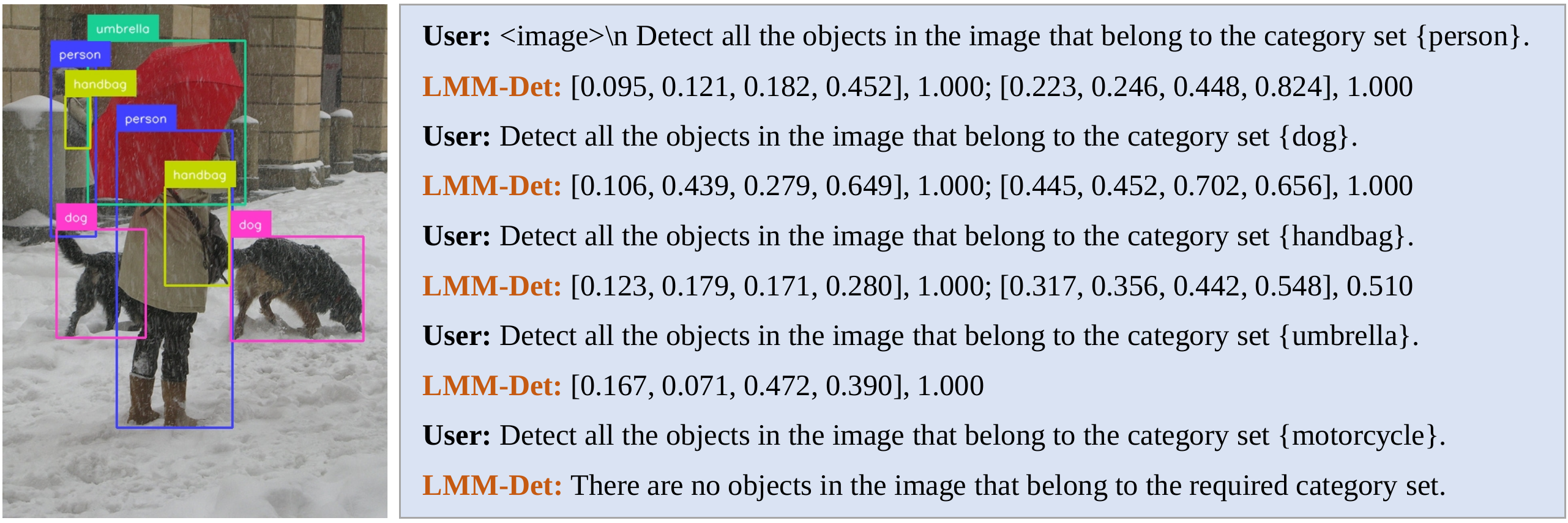}
	\caption{Illustration of the \ljc{re-organization data} designed for object detection over large multimodal models. More details of the prediction confidence of bounding boxes and the post-processing of \xie{LMM-Det's answers} can be referred to Section~\ref{appendix_stat_coco} of Appendix.
	}
	\label{fig:prompt_instruct_dataset}
\end{figure*}

\section{More Qualitative Results.}
\label{appendix_more_qual_res}

The detailed predicted labels of all images in Figure~\ref{fig:first_fig_llava_vs_llava_detection_true_false} of the manuscript are shown in Figure~\ref{appendix_detail_pred_label}.

\ljc{In inference optimization, we re-organize the object detection instruction data to maintain consistency between inference and training. We show an example in Figure~\ref{fig:prompt_instruct_dataset}.}

\begin{figure*}[h]
    \renewcommand\thefigure{C}
    \centering
	\includegraphics[width=0.8\textwidth]{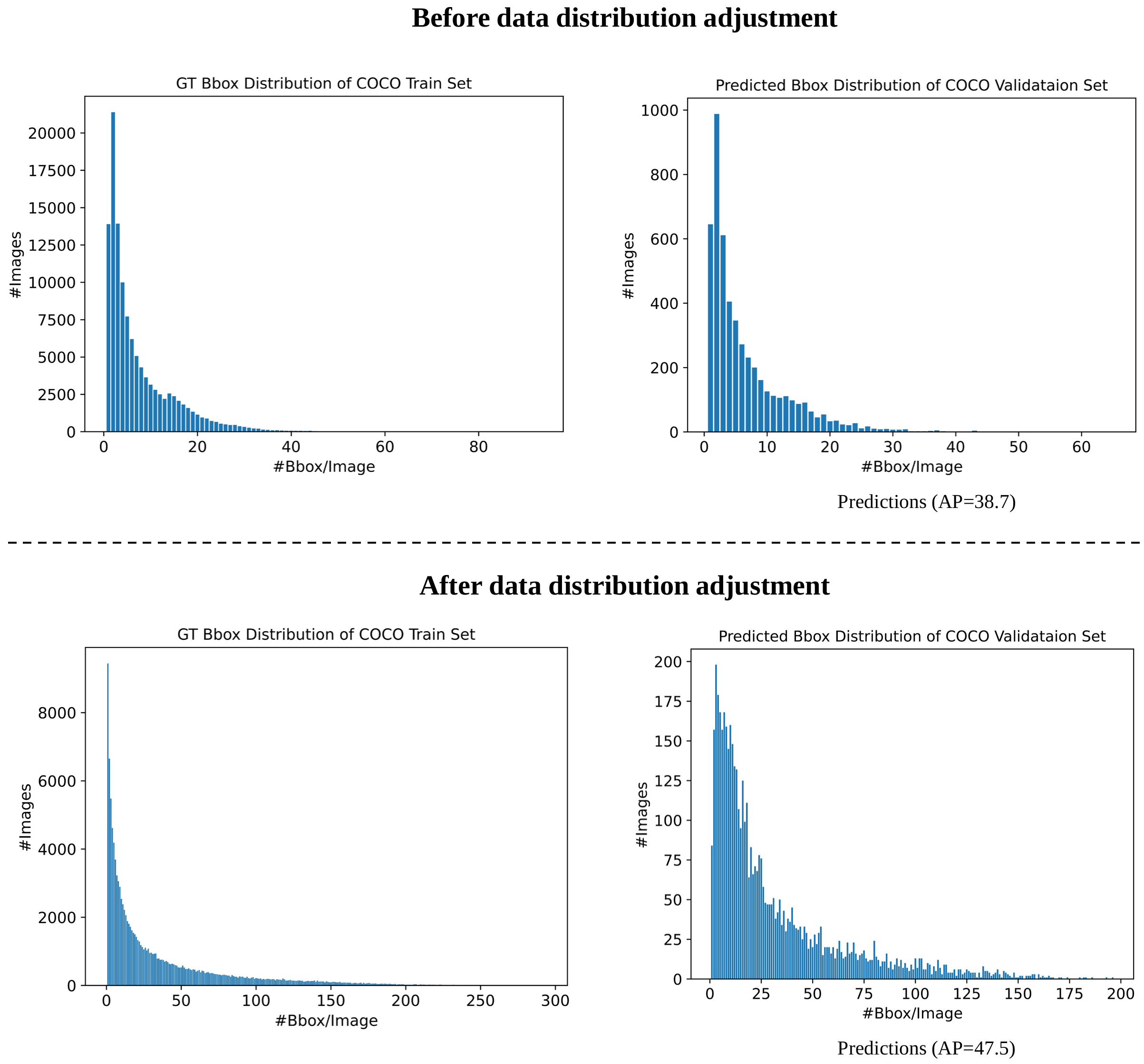}
	\caption{\ljc{Bounding box distribution of ground-truth and predictions before/after data distribution adjustment. The models with AP=38.7 and AP=47.5 can be referred to Table~\ref{finetune}.}
	}
	\label{appendix_adjust_data_distribution}
\end{figure*}

\ljc{We further provide the adjusted bounding box distribution in Figure~\ref{appendix_adjust_data_distribution}. From Figure~\ref{appendix_adjust_data_distribution}, the data distribution adjustment effectively increases the recall rate and improves overall detection performance (the model with AP=47.5).}

\begin{figure*}[t]
    \renewcommand\thefigure{D}
    \centering
	\includegraphics[width=1\textwidth]{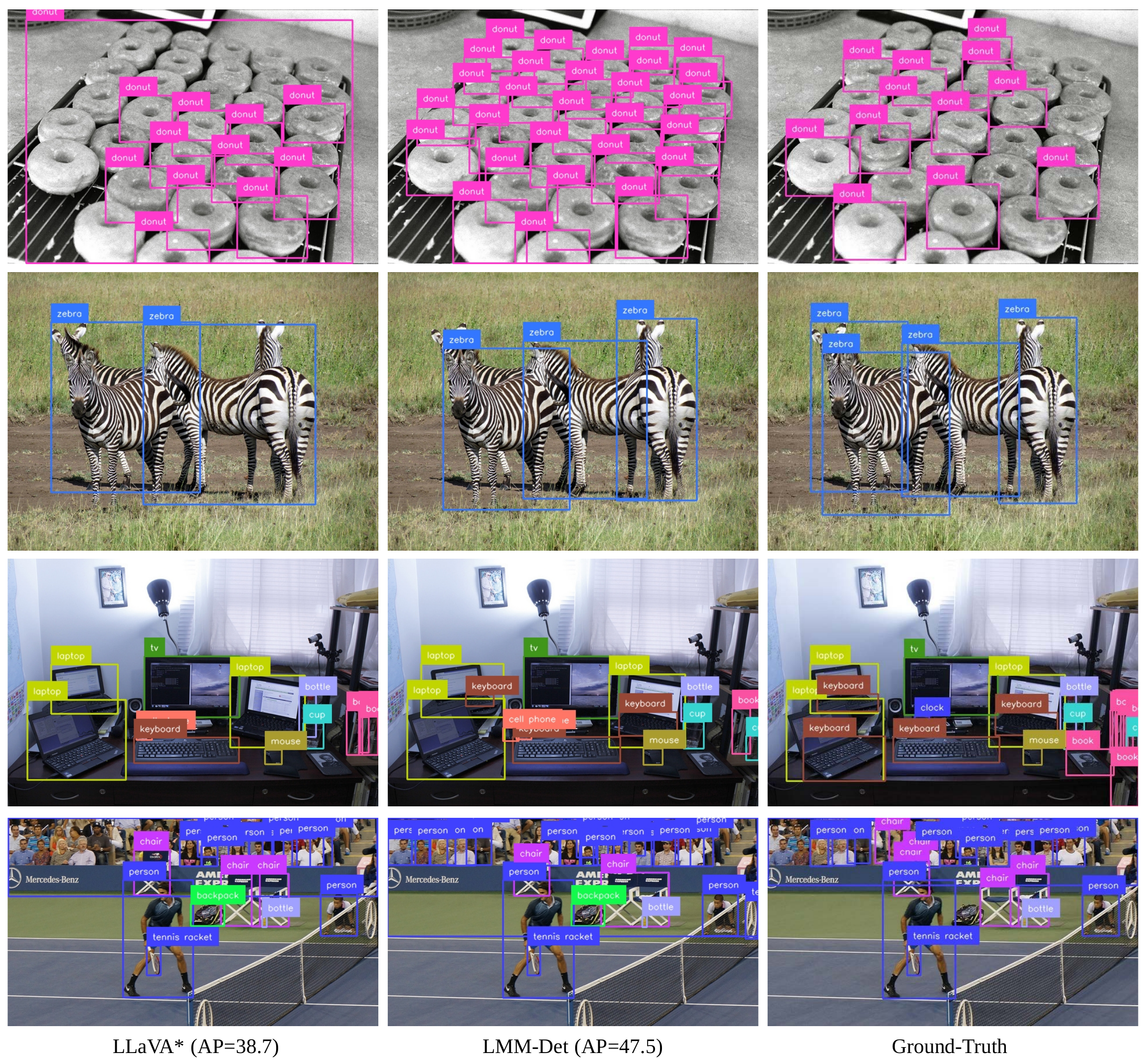}
	\caption{\cam{Qualitative results for recall enhancement.}
	}
	\label{appendix_after_enhancement}
\end{figure*}

As illustrated in the ablation study of the manuscript, the AP of LMM-Det increases from 38.7 to 47.5 after introducing the proposed strategy. We provide the qualitative results for these significant results to show the recall enhancement in Figure~\ref{appendix_after_enhancement}.

\end{document}